\def\year{2022}\relax
\documentclass[letterpaper]{article} 
\usepackage{aaai22}  
\usepackage{times}  
\usepackage{helvet}  
\usepackage{courier}  
\usepackage[hyphens]{url}  
\usepackage{graphicx} 
\usepackage{natbib}  
\usepackage{caption} 
\DeclareCaptionStyle{ruled}{labelfont=normalfont,labelsep=colon,strut=off} 
\usepackage{xcolor}
\usepackage{algorithm}
\usepackage{algorithmic}
\usepackage{newfloat}
\usepackage{listings}
\usepackage{amsmath}
\usepackage{amsthm}
\usepackage{amssymb}
\usepackage{mathrsfs}
\usepackage{dsfont}
\usepackage{bm}
\usepackage{subfig}
\usepackage{booktabs}
\usepackage{diagbox}
\usepackage{multirow}
\usepackage{float}

\floatstyle{ruled}
\newfloat{listing}{tb}{lst}{}
\floatname{listing}{Listing}
\title{Inference-InfoGAN: Inference Independence via Embedding Orthogonal Basis Expansion}
\author{
   Hongxiang Jiang, Jihao Yin, Xiaoyan Luo, Fuxiang Wang
}
\affiliations{
    School of Astronautics, Beihang University \\
    \{jianghongxiang, jihaoyin, luoxy, wangfx\}@buaa.edu.cn

}

\usepackage{bibentry}

\begin{document}

\maketitle

\begin{abstract}
Disentanglement learning aims to construct independent and interpretable latent variables in which generative models are a popular strategy. InfoGAN is a classic method via maximizing Mutual Information (MI) to obtain interpretable latent variables mapped to the target space. However, it did not emphasize independent characteristic. To explicitly infer latent variables with inter-independence, we propose a novel GAN-based disentanglement framework via embedding Orthogonal Basis Expansion (OBE) into InfoGAN network (Inference-InfoGAN) in an unsupervised way. Under the OBE module, one set of orthogonal basis can be adaptively found to expand arbitrary data with independence property. To ensure the target-wise interpretable representation, we add a consistence constraint between the expansion coefficients and latent variables on the base of MI maximization. Additionally, we design an alternating optimization step on the consistence constraint and orthogonal requirement updating, so that the training of Inference-InfoGAN can be more convenient. Finally, experiments validate that our proposed OBE module obtains adaptive orthogonal basis, which can express better independent characteristics than fixed basis expression of Discrete Cosine Transform (DCT). To depict the performance in downstream tasks, we compared with the state-of-the-art Generative Adversarial Network (GAN)-based and even Variational AutoEncoder (VAE)-based approaches on different datasets. Our Inference-InfoGAN achieves higher disentanglement score in terms of FactorVAE, Separated Attribute Predictability (SAP), Mutual Information Gap (MIG) and Variation Predictability (VP) metrics without model fine-tuning. All the experimental results illustrate that our method has inter-independence inference ability because of the OBE module, and provides a good trade-off between it and target-wise interpretability of latent variables via jointing the alternating optimization.
\end{abstract}

\section{Introduction}
Unsupervised representation learning is helpful for downstream tasks like simple classification, reinforcement learning, image translation, spatiotemporal disentanglement and zero-shot learning with insufficient data \cite{bengio2013representation,Chen2016InfoGANIR,don2021pdedriven,Lake2017BuildingMT,wu2019transgaga}. It aims to extract features which are convenient for subsequent processing from unlabeled data. Particularly, disentangled features are more important because of their better independence and interpretability. Therefore, unsupervised disentanglement learning has attracted increasing attention in recent years.

It is natural to use generative model for unsupervised disentanglement learning, because disentanglement can be reflected via observing the noticeable and distinct properties of generated data with the change of one latent variable. It implicitly models the relationship between data and latent variables. The most mainstream generative models in the field of unsupervised disentanglement learning are based on Variational AutoEncoder (VAE) \cite{Kingma2014AutoEncodingVB} or Generative Adversarial Network (GAN) \cite{Goodfellow2014GenerativeAN}. 

Disentanglement learning emphasizes the inter-independence and target-wise interpretability of latent variables \cite{eastwood2018a}. Considering that the VAE objective function focuses on these characteristics, VAE-based disentanglement approaches mostly attempt to modify the objective function form, where $\beta$-VAE \cite{Higgins2017betaVAELB} is a classic work to adjust the weight of independence with an extra hyperparameter $\beta$ in the original VAE objective function. Based on it, subsequent improvements are developed in 
obtaining a Total Correlation (TC) term \cite{5392532} that better constrains independence from independence term \cite{Chen2018IsolatingSO,Jeong2019LearningDA,Kim2018DisentanglingBF}. To solve the uncertainty of model identifiability caused by lacking inductive bias in the above methods, some researchers combine causal inference or nonlinear Independent Component Analysis (ICA) \cite{khemakhem2020variational,yang2020causalvae}. In general, these methods have strong disentanglement ability, but tend to generate low-quality images. To improve the quality of the generated images, GAN-based approaches are proposed. For example, InfoGAN \cite{Chen2016InfoGANIR} generates images correlated with latent variables by maximizing the Mutual Information (MI) between them. Although it can generate clear images, the performance on disentanglement is not enough, because the independence of latent variables is ignored. Then, many approaches try to embed contrastive learning \cite{Deng2020DisentangledAC,Lin2020InfoGANCRAM,Zhu2020LearningDR}, VAEs \cite{NIPS2016_ef0917ea}, and Conditional Batch Normalization (CBN) \cite{NIPS2017_6fab6e3a,miyato2018cgans,nguyen2019hologan} to constrain independent latent variables, in which InfoGAN-CR performs well \cite{Lin2020InfoGANCRAM}, even far exceeding VAE-based counterparts. To control the independence of latent variables, it uses contrastive latent sampling as the introduction of prior knowledge, indirectly exploiting self-supervised information.

In this work, we focus our study on building a GAN-based model with inter-independent inference ability between latent variables. To infer inter-independence on keeping target-wise interpretability, a novel Orthogonal Basis Expansion (OBE) module is proposed and embedded into the GAN-based model, named Inference-InfoGAN, which directly employs self-supervised information of the data without introducing more priors. Our main contributions are as the following three aspects.

\begin{itemize}

\item We introduce an Inference-InfoGAN framework, embedding the designed OBE module into InfoGAN structure. Owing to the OBE, an adaptive orthogonal basis can be obtained to represent the data and infer the inter-independence between latent variables. Following the InfoGAN structure, the target-wise interpretability remains similar to the traditional GAN-based methods.

\item We design an alternating optimization on consistence constrain and orthogonal requirement updating. This step can balance both abilities between the inter-independent inference and target-wise interpretable consistency on training of Inference-InfoGAN. 

\item We compare our proposed method with the state-of-the-art GAN-based and VAE-based models, aided by downstream tasks. The disentanglement performance is evaluated on unlabeled datasets by unsupervised disentanglement metrics, besides evaluating on labeled datasets. The experimental results illustrate the higher disentanglement to mine richer self-supervised information without specific supervision information.

\end{itemize}

\section{Related Work}

\subsection{InfoGAN-based models}
InfoGAN is a model combined with GAN and information theory. Based on GAN, a generator $G$ and a discriminator $D$ can be trained by the idea of game theory. When the optimal solutions are obtained, the $G$ can map the noise ${\rm\textbf{z}}$ to the real data distribution ${\rm\textbf{x}}\sim p_{data}({\rm\textbf{x}})$. It makes that $D$ cannot distinguish synthetic data from real data. The objective function of GAN is described as a minimax game as follow:
\begin{equation}
\label{eq:GAN loss}
\begin{aligned}
    \mathop{\min}\limits_{G}\mathop{\max}\limits_{D}\mathcal{L}_{adv}(D,G)= \mathbb{E}_{{\rm\textbf{x}}\sim p_{data}({\rm\textbf{x}})}\left[\log{D({\rm\textbf{x}})}\right] \\ +\mathbb{E}_{{\rm\textbf{z}}\sim p_{z}({\rm\textbf{z}})}\left[\log{(1-D(G({\rm\textbf{z}})))}\right],
\end{aligned}
\end{equation}
where $p_{z}({\rm\textbf{z}})$ represents a standard Gaussian distribution. Eq. \ref{eq:GAN loss} can be transformed into minimizing the Jensen-Shannon (JS) divergence of the real data distribution $p_{data}$ and the generated data distribution $p_{G}$. Subsequent research on stabilizing its training and improving its performance tries to introduce Wasserstein Distance in the optimal transmission theory \cite{villani2009optimal} to replace JS divergence \cite{Arjovsky2017WassersteinGA,Gulrajani2017ImprovedTO,Wu2018WassersteinDF}. Furthermore, SNGAN \cite{miyato2018spectral} uses spectral normalization to ensure a Lipschitz constraint which is necessary to optimize the Wasserstein Distance.

Generally, vanilla GAN is a breakthrough work to generate high-quality images, but the generation is random without controlling. To guide the generation of images with specific interpretable characteristics, InfoGAN is developed. It applies GAN to the field of unsupervised disentanglement learning for the first time, and is competitive with supervised models. InfoGAN achieves interpretable feature representation by maximizing the MI between the partial latent variables ${\rm\textbf{c}}$ and the images ${{\rm\textbf{x}}=G({\rm\textbf{z}},{\rm\textbf{c}})}$. It proposes the following information-regularized minimax game:

\begin{equation}
\label{eq:infoGAN loss}
\begin{aligned}
    \mathop{\min}\limits_{G,Q}\mathop{\max}\limits_{D}\mathbf{}\mathcal{L}_{InfoGAN}(D,G,Q)= \mathcal{L}_{adv}(D,G) \\ -\lambda{\mathcal{L}_{I}(G,Q)}.
\end{aligned}
\end{equation}

This method defines a variational lower bound of the MI:
\begin{equation}
\label{eq:lbmi loss}
\begin{aligned}
    \mathcal{L}_{I}(G,Q)&=\mathbb{E}_{{\rm\textbf{c}}\sim p({\rm\textbf{c}}),{\rm\textbf{x}}\sim p(G({\rm\textbf{z}},{\rm\textbf{c}}))}\left[\log{Q({\rm\textbf{c}}|{\rm\textbf{x}})}\right]+H({\rm\textbf{c}})\\
    &\leq {I}({\rm\textbf{c}};G({\rm\textbf{z}},{\rm\textbf{c}})),
\end{aligned}
\end{equation}
and uses the neural network to fit the variational distribution so that MI can be estimated and maximized. However, it is insufficient in inter-independence characteristics only maximizing the MI. Recently, some advanced methods try to improve InfoGAN. For example, IB-GAN \cite{Jeon2018IBGANDR} is inspired by Information Bottleneck (IB) theory \cite{ShwartzZiv2017OpeningTB} via adding a capacity regularization for MI directly. OOGAN \cite{liu2020oogan} adds orthogonal regularization with constraining the network weight of $Q$. InfoGAN-CR \cite{Lin2020InfoGANCRAM} introduces contrastive learning to provide additional prior knowledge and exploit self-supervised information indirectly.

\subsection{$\beta$-VAE-based models}
$\beta$-VAE simply changes the objective function of VAE to:
\begin{equation}
\label{eq:beta-VAE loss}
\begin{aligned}
    \mathbb{E}_{q_\phi({\rm\textbf{z}|\textbf{x}})}\left[\log{p_\theta({\rm\textbf{x}|\textbf{z}})}\right]-\beta{D_{KL}}\left(q_\phi({\rm\textbf{z}|\textbf{x}})||p({\rm\textbf{z}})\right),
\end{aligned}
\end{equation}
where ${\rm\textbf{z}}$ denotes the latent variables that control the generated images ${\rm\textbf{x}}$, $p_\theta$ and $q_\phi$ are fitted by two neural networks. $\beta$-VAE maximizes the Evidence Lower-Bound (ELBO) while setting $\beta>1$ to increase the penalty for the $D_{KL}\left(q_\phi({\rm\textbf{z}|\textbf{x}})||p({\rm\textbf{z}})\right)$ term ($\beta=1$ in the vanilla VAE). When $p({\rm\textbf{z}})$ is defined as an independent Gaussian prior, the larger $\beta$, the more $q_\phi({\rm\textbf{z}|\textbf{x}})$ tends to an independent distribution, which enhances independent disentangled representation and reduces the quality of image reconstruction. 

The following research tries to introduce a variational distribution again, and separates a new TC regularization item from $D_{KL}$ term:
\begin{equation}
\label{eq:beta-TCVAE TC term}
\begin{aligned}
    \mathcal{L}_{TC} = D_{KL}(q({\rm\textbf{z}})||\prod_j{q({\rm\textbf{z}}_j)}),
\end{aligned}
\end{equation}
to directly constrains the independence of latent variables. However, because the TC term is intractable, to optimize the objective function, $\beta$-TCVAE \cite{Chen2018IsolatingSO} uses an approximate method of estimating TC, while FactorVAE \cite{Kim2018DisentanglingBF} estimates the distribution via adding a discriminator.

\section{Method}
\label{sec:Method}

InfoGAN can generate high-quality images from controlled latent variables with less independence, while $\beta$-VAE can construct high-independence latent variables with lower-quality images. To integrate the goal of high-quality images in InfoGAN and high-independence latent variables in $\beta$-VAE, we propose a novel independent inference framework based on InfoGAN in this paper. 

Following the generative model, real image data $\bm{X}^*\in{\mathbb{R}^{n\times{n}}}$ comes from a continuous control vector $\bm{c}\in{\mathbb{R}^{k}}$ and uncontrollable noise $\bm{z}\in{\mathbb{R}^{d}}$, namely there is an optimal mapping $G^*$ that satisfies $\bm{X}^*=G^*(\bm{z},\bm{c})$. A discriminator $D$ distinguishes the distribution of real and generated data. To facilitate the search for the optimal mapping stably, the adversarial loss $\mathcal{L}_{adv}(D,G)$ is applied the same as in the WGAN-div:
\begin{equation}
\label{eq:WGAN-div loss}
\begin{aligned}
    \mathop{\min}\limits_{G}\mathop{\max}\limits_{D}\mathcal{L}_{adv}(D,G)= \mathbb{E}_{{\bm{X}}\sim p_{data}({\bm{X}})}\left[D({\bm{X}})\right] \\ -\mathbb{E}_{{\bm{z}}\sim p_{z}({\bm{z}})}\left[D(G({\bm{z}}))\right] -k\mathbb{E}_{\hat{\bm{X}}\sim p_{u}({\hat{\bm{X}}})}[{\Vert {\nabla}_{\hat{\bm{X}}} D(\hat{\bm{X}}) \Vert}],
\end{aligned}
\end{equation}
where $\hat{\bm{X}}$ comes from a linear combination of real and generated data. Based on these denotations, an unsupervised disentanglement problem can be modelled as each dimension in $\bm{c}$ independently control an interpretable feature of $\bm{X}^*$. To promote the expression of independent characteristics, we conduct latent variables inference on the generated image $\bm{X}$ based on its adaptive orthogonal basis which can be obtained from our specially designed OBE module. Additionally, we propose an integrated Inference-MI loss $\mathcal{L}_{infer-info}(G,Q,\bm{P})$, where $Q$ is an encoder based on InfoGAN, and $\bm{P}\in{\mathbb{R}^{n\times{n}}}$ is an adaptively obtained orthogonal matrix to further restrict the inter-independence of $\bm{c}$. In order to ensure the orthogonality of $\bm{P}$, we introduce the method of orthogonal regularization. Orthogonal regularization often appears in the constraints on network weight matrix $\bm{W}$, which satisfies $\bm{WW}^T-\bm{I}=\bm{0}$. Most approaches construct the loss function based on this. For example, Brock \cite{L1Orth} achieves goals by minimizing the following formula: 
\begin{equation}
\label{eq:L1}
\begin{aligned}
    \sum(|\bm{WW}^T-\bm{I}|).
\end{aligned}
\end{equation}

BigGAN \cite{Brock2019LargeSG} optimize by minimizing the Frobenius norm of the matrix, and SNGAN iteratively calculates the 2-norm of the matrix \cite{miyato2018spectral}. In addition, there are methods based on Singular Value Decomposition (SVD) to satisfy orthogonal constraints \cite{Li2021OrthogonalDN}. To quickly find an orthogonal matrix $\bm{P}$, similar to Eq .\ref{eq:L1}, we minimize the following new objective function:

\begin{equation}
\label{eq:Lor}
\begin{aligned}
    \mathcal{L}_{or}=\sum(|\bm{PP}^T-\bm{I}|).
\end{aligned}
\end{equation}

Based on the orthogonal matrix $\bm{P}$, we can perform disentanglement-related inference on the generated image. Generally, for data $\bm{X}$ in a finite dimensional space, it can be expanded as follows:
\begin{equation}
\label{eq:obe}
\begin{aligned}
    \bm{X}=\sum_{m=1}^Mc'_m\bm{e_m},
\end{aligned}
\end{equation}
where $\bm{e_m}\in{\mathbb{R}^{n\times{n}}}$, ${\left\{\bm{e_m}\right\}}_{m=1}^M$ denotes the standard orthogonal basis, ${M}$ denotes the number of the basis. Using the elements in $\bm{P}$, the orthonormal basis ${\left\{\bm{e_m}\right\}}_{m=1}^M$ and the result of the expansion of $\bm X$ will be explicitly expressed as: 
\begin{equation}
\label{eq:search x}
\begin{aligned}
    \bm{X}&=\bm{PC}'\bm{P}^T\\&=\sum_{i=1}^n\sum_{j=1}^nc'_{ij}\bm{p_i}\bm{p}_{\bm{j}}^T\notag,
\end{aligned}
\end{equation}
where ${\bm{p_i}\in{\mathbb{R}^{n\times{1}}}}$, ${\bm{C}'\in{\mathbb{R}^{n\times{n}}}}$. Thus, when ${i_1}={i_2}$ and ${j_1}={j_2}$, Hadamard product of ${\bm{p_{i_1}}}\bm{p}_{\bm{j_1}}^T$ and ${\bm{p_{i_2}}}\bm{p}_{\bm{j_2}}^T$ is $1$, otherwise it is $0$, i.e. the standard orthogonal basis has been found, which is ${\left\{\bm{p_1}\bm{p}_{\bm{1}}^T, \bm{p_1}\bm{p}_{\bm{2}}^T, ..., \bm{p_1}\bm{p}_{\bm{n}}^T, \bm{p_2}\bm{p}_{\bm{1}}^T, ..., \bm{p_n}\bm{p}_{\bm{n}}^T\right\}}$. The coefficients corresponding to different basis could have an independent and strong correlation effect on $\bm{X}$. Therefore, excellent orthogonal matrix $\bm{P}$, generator $G$, and encoder $Q$ jointly achieve the following effects: If the controllable latent variables $\bm{c}$ is changed, the specific features of mapped image under $G$ can be changed. It is manifested in the MI increasing estimated by $Q$ between the images and the latent variables, and the coefficient changing represented by the OBE inference on images. 

Accordingly, we design the following maximization goal function:

\begin{equation}
\label{eq:9}
\begin{aligned}
    &{I}(\bm{c};G(\bm{z},\bm{c})) \\ -&\lambda\mathbb{E}_{{\bm{X}}\sim p(G(\bm{z},\bm{c}))}\left[D_{KL}\left(p(\bm{c}|\bm{X})||\prod_i q'(c_i|\bm{X};\bm{P})\right)\right].
\end{aligned}
\end{equation}

Therefore, maximizing Eq. \ref{eq:9} is to maximize the first mutual information item and minimize a Kullback-Leibler (KL) divergence, which can push the unknown and real conditional distribution $p(\bm{c}|\bm{X})$ to an independent distribution $\prod_i q'(c_i|\bm{X};P)$, this distribution can be expressed as following as follows with Eq. \ref{eq:search x}:
\begin{figure*}[htb]
\centering
\includegraphics[width=0.8\textwidth]{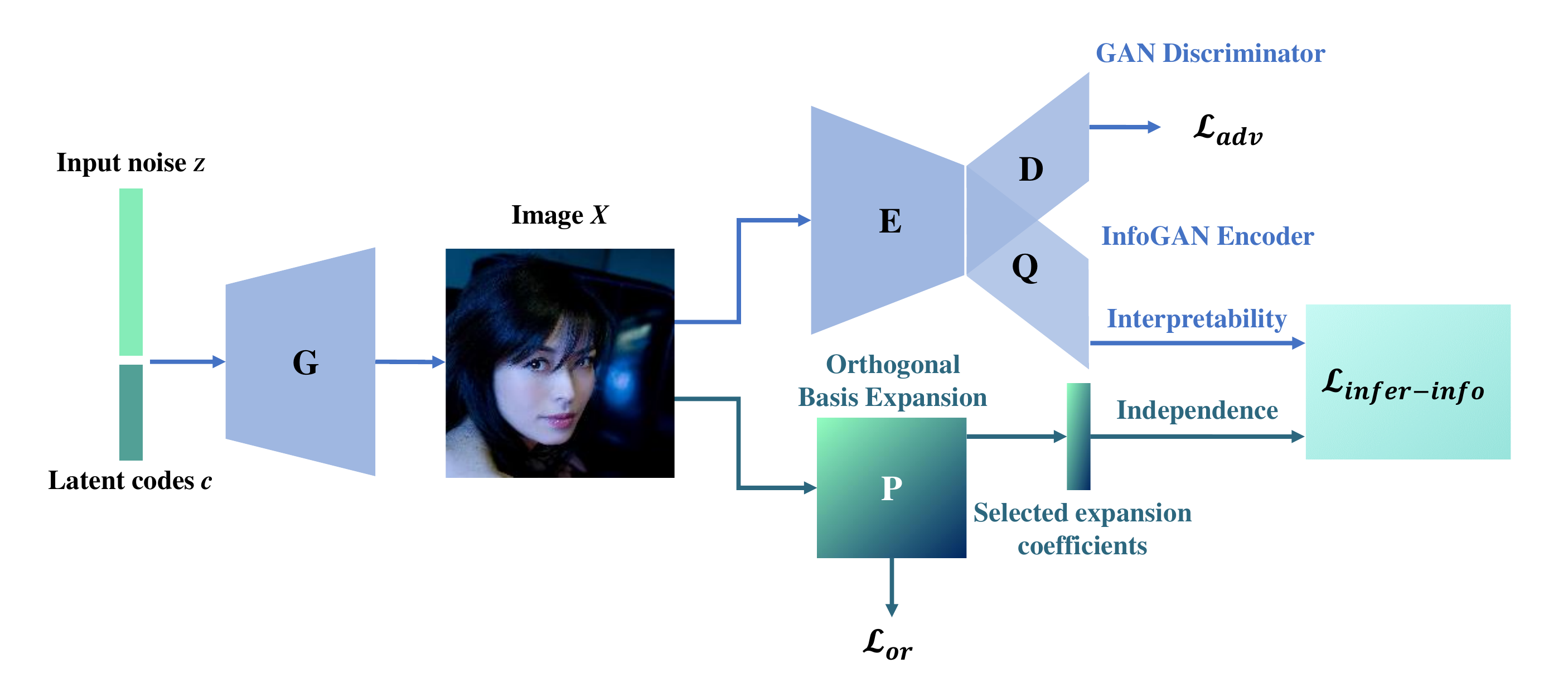}
\caption{\textbf{Inference-InfoGAN module.} InfoGAN only includes the blue part of the model, Inference-InfoGAN adds the branch below, where $\bm{P}$ represents the parameters in our proposed OBE module, adaptively obtaining orthogonal basis and inferring latent variables with inter-independence.}
\end{figure*}
\begin{equation}
\label{eq:10}
\begin{aligned}
    \prod_i q'(c_i|\bm{X};\bm{P})&=\prod_i q'(c_i|\bm{PC}'\bm{P}^T;\bm{P})\\
    &=\prod_i q'(c_i|c'_1, c'_2, ...;\bm{P})\\
    &=\prod_i q'(c_i|c'_i;\bm{P}),
\end{aligned}
\end{equation}
where $c'_i$ is the element in $\bm{C}'=\bm{P}^T\bm{XP}$. In order to optimize Eq. \ref{eq:9}, we further use the technique of Variational Information Maximization to derive its lower bound, $\mathcal{L}_{infer-info}(G,Q,\boldsymbol{P})$, similar to InfoGAN \cite{Barber2003TheIA}.The detailed proof is shown in Appendix A:
\begin{equation}
\label{eq:inter-info}
\begin{aligned}
    &\mathcal{L}_{infer-info}(G,Q,\bm{P}) \\ 
    =&\mathbb{E}_{\bm{c}\sim p(\bm{c}),\bm{X}\sim p(G(\bm{z},\bm{c}))}\bigg[\lambda\log{\prod_iq'(c_i|c'_i;\bm{P})}\\
    +&(1-\lambda)\log{q(\bm{c}|\bm{X};Q)}\bigg],
\end{aligned}
\end{equation}
where hyperparameter $\lambda\in(0,1)$ affects the weight of the two items in $\mathcal{L}_{infer-info}(G,Q,\boldsymbol{P})$, the first one is used to constrain inter-independence and consistence between the OBE-based inference results and latent variables, the second one is used to emphasize target-wise representation. The larger $\lambda$, the greater the penalty for the independence of latent variables. The smaller $\lambda$ reflects the more similarity to the vanilla InfoGAN, resulting in poor disentanglement performance. We can simply optimize $\mathcal{L}_{infer-info}(G,Q,\boldsymbol{P})$ using Monte Carlo method. 

Overall, our total objective function can be composed of the adversarial loss $\mathcal{L}_{adv}(D,G)$, the Inference-MI loss $\mathcal{L}_{infer-info}(G,Q,\boldsymbol{P})$ and the orthogonality loss $L_{or}$ with non-negative scalar $\alpha$ and $\gamma$: 

\begin{equation}
\label{eq:Inference-InfoGAN loss}
\begin{aligned}
    \mathop{\min}\limits_{G,Q,\boldsymbol{P}}\mathop{\max}\limits_{D}\mathcal{L}(D,G,Q,\boldsymbol{P})=\mathcal{L}_{adv}(D,G) \\ -\gamma\mathcal{L}_{infer-info}(G,Q,\boldsymbol{P})+\alpha\mathcal{L}_{or}(\boldsymbol{P}).
\end{aligned}
\end{equation}

Based on the above method, we build the model in Figure 1.

In fact, $L_{or}$ only calculates the gradient of the parameters in the OBE module. Other parts optimize all parameters, constraining the interpretability of the generated image and the consistence with the latent variables. In order to balance the training of the orthogonal basis and consistence constrain, we design an alternating optimization step: After updating $G$, $D$, $Q$ and $\boldsymbol{P}$ according to $\mathcal{L}_{adv}(D,G)$ and $\mathcal{L}_{infer-info}(G,Q,\boldsymbol{P})$ in each iteration, we fix the parameters in $G$, $D$, $Q$, and only repeatedly train $\bm{P}$ according to $L_{or}$. When $L_{or}$ is below an upper bound $\epsilon$, proceed to the next iteration.

This method balances the orthogonality of the basis and other constraints. Although the training speed of each epoch is slightly reduced, it is more convenient to find a model that performs well. The pseudocode of the entire model training is given in Appendix B.

\section{Experiment}

\subsection{Datasets and evaluation metrics}

To demonstrate the performance of our method, we conduct some qualitative and quantitative experiments on two traditional datasets, including the dSprites \cite{dsprites17} dataset with available disentangling annotations and the CelebA \cite{liu2015faceattributes} dataset without the corresponding disentanglement labels. For quantitative evaluation, we adopt three evaluation metrics on the dSprites dataset, namely the FactorVAE, Separated Attribute Predictability (SAP), and Mutual Information Gap (MIG) \cite{Chen2018IsolatingSO,Kim2018DisentanglingBF, kumar2018variational}. On the other hand, inspired by \cite{locatello2019challenging}, for a fair comparison to show that our work has stable and superior performance on unlabeled datasets without specific supervision information for model fine-tuning, two metrics are used for the CelebA dataset, i.e. the Variation Predictability (VP) disentanglement metric \cite{Zhu2020LearningDR} and the image quality metric Fréchet Inception Distance (FID) \cite{Heusel2017GANsTB}.

In fact, VP score is calculated by predicting which dimension of the variable has changed from a pair of images, which are the results generated by the generator $G$ trained by the disentanglement model and only one-dimensional latent variable in the input of $G$ is different. Therefore, VP score reflects the performance in downstream tasks of classifying or cluster different features \cite{locatello2020weakly}. It should be noted that downstream tasks on the CelebA dataset should not use our model to generate false information to deceive humans, even though it has such a potential possibility.

We use all the datasets and reference codes in accordance with their licenses. The source code of calculating metrics by Locatello et al. \cite{locatello2019challenging} uses the Apache license, FactorVAE uses the MIT license, the dSprites dataset uses the Apache license, and the CelebA dataset can be used for non-commercial research or educational purposes.

\subsection{Implementation details}

All models in the experiments are optimized with batchsize set to $64$, using Adam optimizer \cite{DBLP:journals/corr/KingmaB14} with initial learning rate set to $0.0009$, $\beta_1$ set to $0.5$ and $\beta_2$ set to $0.999$. The baseline model used in our experiments introduces the Wasserstein Distance and zero-mean penalty to InfoGAN, named InfoWGAN-GP (modified). In other words, InfoWGAN-GP (modified) just removes the OBE module in our models. For the dSprites datasets, we set the size of the generated image to $64$ $\times$ $64$, and control the dimension of the latent variables $\bm{c}$ to $5$, the dimension of the noise $\bm{z}$ to $60$. The optimal hyperparameter settings are $\lambda=0.9$, $\gamma=1.1$ $\epsilon=0.2$. For the CelebA datasets, we set the size of the generated image to $128$ $\times$ $128$, and control the dimension of the latent variables $\bm{c}$ to $15$, the dimension of the noise $\bm{z}$ to $120$. The hyperparameter $\epsilon=0.8$, others are the same as the optimal settings on the dSprites datasets. For a multi-channel image, after OBE for each channel, there are multiple expansion coefficients on the same basis, so we add a layer of convolution to combine the multiple coefficients into one. When calculating the VP score, all models set the training set ratio to $0.1$, epoch to $200$, and the highest test result is selected. When calculating the FID, the number of selected images is $128000$, which can get a reliable result. All the experiments are carried out on a server with NVIDIA $1080$ Ti GPUs.

\subsection{Experiments on dSprites datasets}
\label{Experiments on dSprites datasets}

We compare the proposed method with four traditional VAE-based methods (beta-VAE, FactorVAE, CasVAE and CasVAE-VP) and four GAN-based methods (InfoGAN, InfoWGAN-GP (modified), InfoGAN-CR and InfoGAN-CR (model selection)). In addition, the Inference-InfoGAN (DCT) is also evaluated in this experiment, which alternatively adopts Discrete Cosine Transform (DCT) for inference based on the OBE module (seeing Appendix C). 

\begin{figure}[htb]
	\centering
	\subfloat[Latent traversals trained on dSprites]{\includegraphics[width=0.4\textwidth]{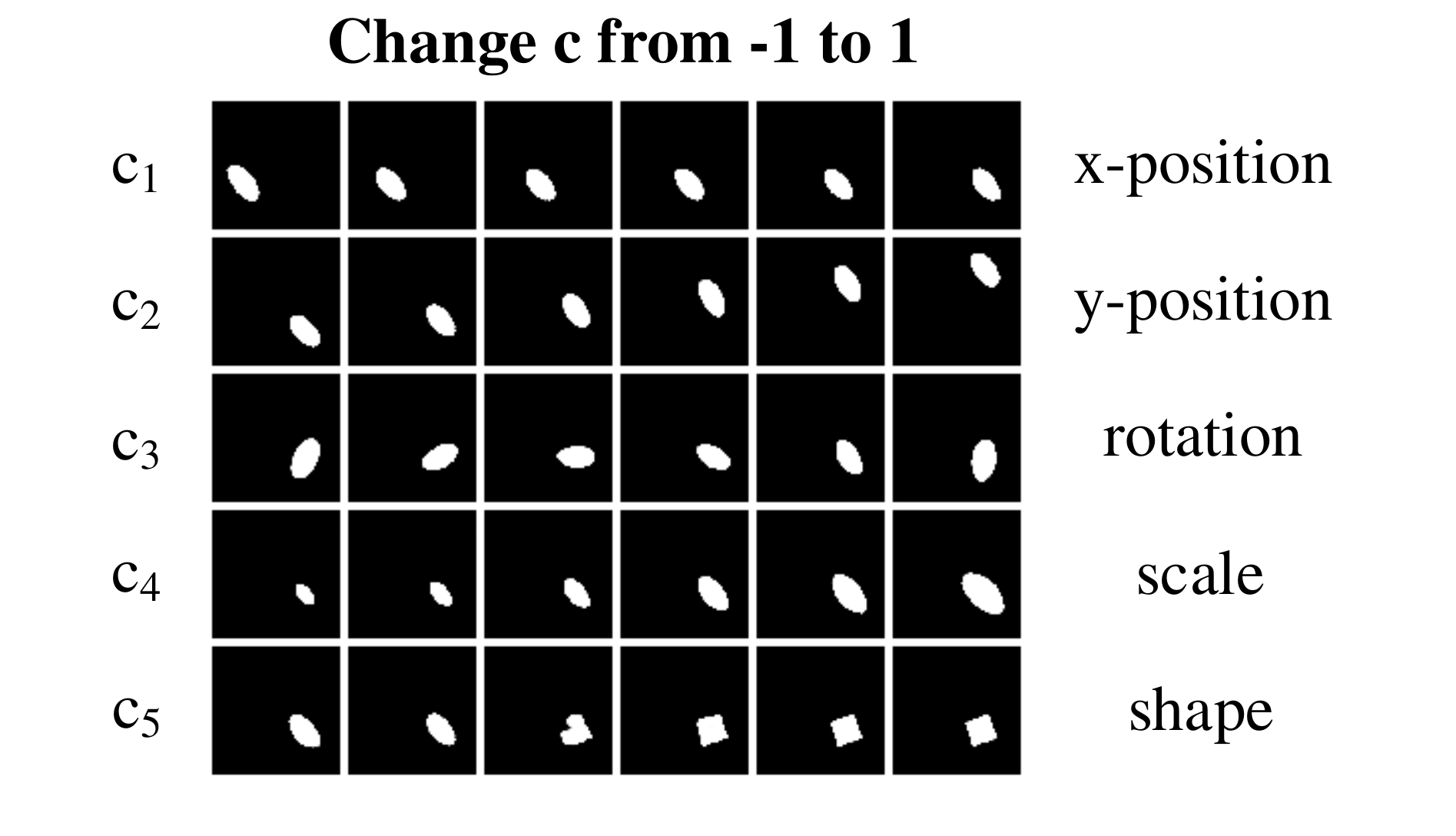}}
	\newline
	\subfloat[FactorVAE disentanglement score]{\includegraphics[width=0.4\textwidth]{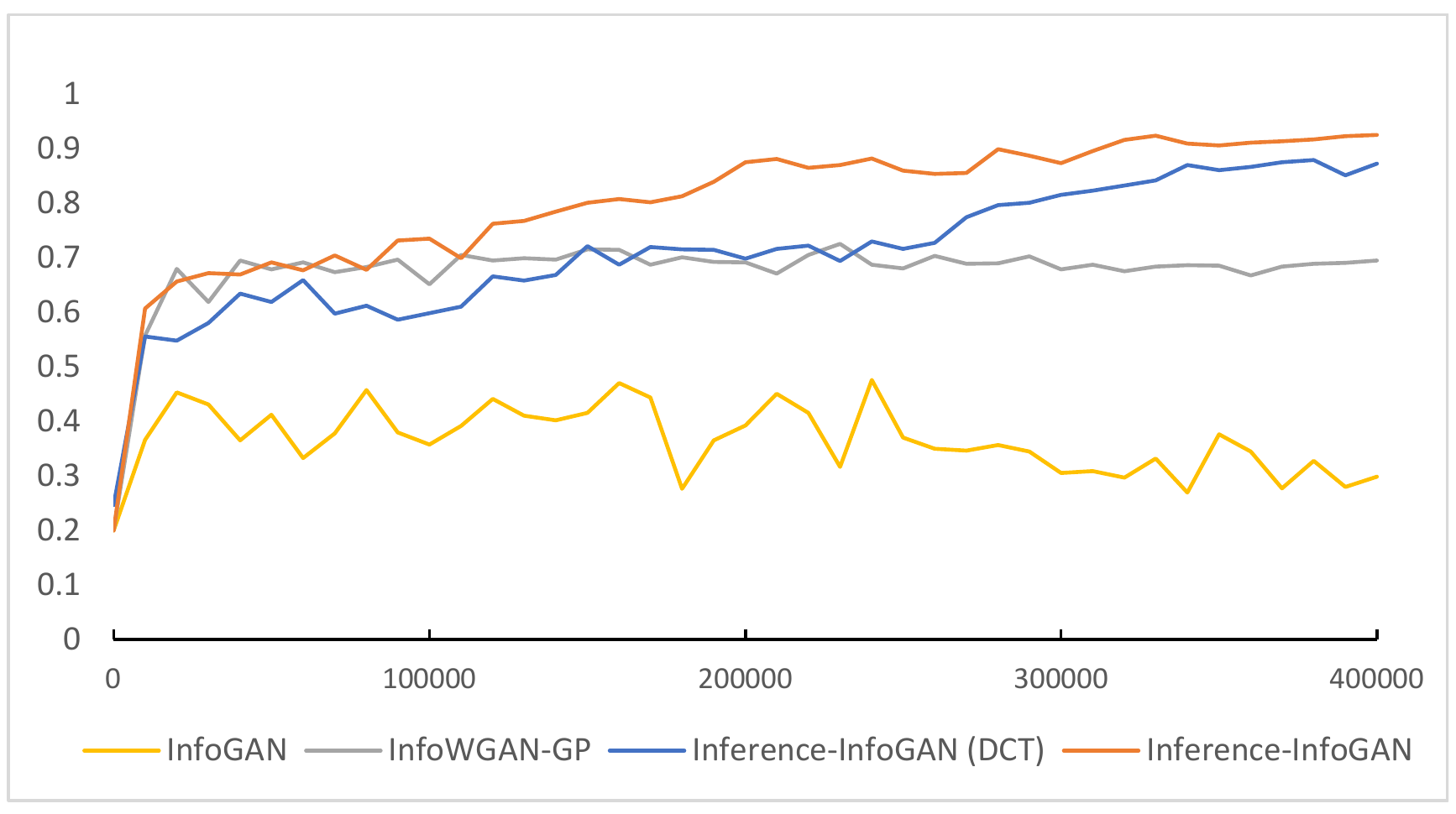}}
	\caption{\textbf{Experimental results on the dSprites dataset}. (a) Only by changing the value of the latent variables $\bm{c}$ in a single dimension, the corresponding generated images will only change in a certain feature, which is also consistent with the dataset and has strong interpretability. (b) The curve shows the FactorVAE score for different methods. }
	\label{fig:figure dsprites}
\end{figure}

For qualitative analysis, the latent traversals obtained by our proposed Inference-InfoGAN is shown in Figure \ref{fig:figure dsprites}(a). It can be observed that our method expresses interpretable information sufficiently, i.e. horizontal and vertical directions, rotation, scale and shape. 
\begin{table*}[htb]
    \caption{\textbf{Disentanglement scores on dSprites dataset}. The range of disentanglement scores is $[0,1]$ and $1$ is the best. To enhance the robustness of the results, we only change the random seed and train our Inference-InfoGAN models in $8$ times. On FactorVAE metric, no matter how we choose orthogonal basis in our method, the result is higher than other advanced work based on GAN or VAE. On SAP and MIG metric, our method is slightly better than InfoGAN-CR.}
	\centering
	\setlength{\tabcolsep}{1.3mm}{
		\linespread{1.1}\selectfont
		\begin{tabular}{cccc}  \toprule
			Model  &FactorVAE &SAP &MIG \\  \hline
			beta-VAE  &$0.630\pm0.100^\dagger$ &$0.550\pm0.000^\dagger$ &$0.210\pm0.000^\dagger$\\
			FactorVAE  &$0.840\pm0.000^\dagger$ &$0.580\pm0.000^\dagger$ &$0.150\pm0.000^\dagger$\\
			CasVAE  &$0.913\pm0.074^\ddagger$ & \\
			CasVAE-VP  &$0.917\pm0.069^\ddagger$ & \\
			\hline
			InfoGAN   &$0.476\pm0.012$ &$0.078\pm0.000$ &$0.019\pm0.000$\\
			InfoWGAN-GP (modified)  &$0.771\pm0.012$ &$0.449\pm0.001$ &$0.182\pm0.000$\\
			InfoGAN-CR  &$0.880\pm0.010^\dagger$ &$0.580\pm0.010^\dagger$ &$0.370\pm0.010^\dagger$\\
			InfoGAN-CR (model selection)  &$0.920\pm0.000^\dagger$ &$\textbf{0.650}\pm\textbf{0.000}^\dagger$ &$\textbf{0.450}\pm\textbf{0.000}^\dagger$\\
			\hline
			Inference-InfoGAN (DCT)  &$\textbf{0.927}\pm\textbf{0.008}$ &$0.598\pm0.001$ &$0.385\pm0.001$\\
			Inference-InfoGAN ($\lambda=0.9$, $\gamma=1.1$)  &$\textbf{0.946}\pm\textbf{0.015}$ &$0.622\pm0.023$ &$0.408\pm0.025$\\
			Inference-InfoGAN ($\lambda=0.5$, $\gamma=1$)  
			&$0.884\pm0.035$ &$0.518\pm0.018$ &$0.226\pm0.033$\\
			Inference-InfoGAN ($\lambda=0.1$, $\gamma=1$)  
			&$0.735\pm0.076$ &$0.394\pm0.046$ &$0.171\pm0.039$\\
			\hline
			Best model in InfoGAN-CR (model selection)	
			&$0.950^\dagger$ &$0.650^\dagger$ &$\textbf{0.460}^\dagger$\\
			Best model (ours)	
			&$\textbf{0.966}$ &$\textbf{0.658}$ &$0.453$\\
			\hline
	\end{tabular}}
\label{tab:Disentanglement scores dsprites}
\end{table*}

The quantitative disentanglement results are shown in Figure \ref{fig:figure dsprites}(b) (FactorVAE disentanglement score curves) and Table \ref{tab:Disentanglement scores dsprites}, where the results with $\dagger$ and $\ddagger$ are directly obtained from the paper \cite{Lin2020InfoGANCRAM} and \cite{Zhu2020LearningDR} respectively. Our model is steadily optimized as the training iteration increases, and eventually achieves the best performance on FactorVAE among all the models in Table \ref{tab:Disentanglement scores dsprites}. Specifically, our proposed method enhances by about $0.175$ Factor-VAE, $0.173$ SAP and $0.226$ MIG compared to the baseline InfoWGAN-GP. Furthermore, it is higher than InfoGAN-CR, the state-of-the-art model in recent one year, on all metrics ($0.066$ on FactorVAE, $0.042$ on SAP, $0.038$ on MIG). Even if InfoGAN-CR employs model selection of ModelCentrality, our FactorVAE score is still $0.026$ higher than it. In addition, our best model is better than its best model of InfoGAN-CR (model selection) on FactorVAE and SAP, only slightly worse on MIG. Therefore, our work is competitive to disentanglement learning of InfoGAN-CR based on GAN. It focuses on improving the independence of latent variables via seeking an adaptive orthogonal basis without adding any model selection technique.
In particular, Inference-InfoGAN obtains better results than the DCT version. Thus, it can be observed that the good performance greatly benefits from the proposed OBE module considering interpretability and independence for the disentanglement task and adaptive orthogonal basis which has better ability to express independent characteristics than fixed DCT strategy.

It should be noted that our OBE module, alternating optimization, and the InfoGAN loss are all effective and critical components of the network. We find that when removing the vanilla InfoGAN loss or the OBE module, the disentanglement ability of the model will be greatly reduced. After replacing alternating optimization with one-step training, albeit the training process is more stable and the disentanglement ability evaluated by FactorVAE remains high, model cannot have the same high level of performance on other metrics. The detailed experimental evaluation will be given in the ablation studies section.

\subsection{Experiments on CelebA datasets}

We compare the proposed method with one traditional VAE-based method (FactorVAE) and four GAN-based methods (InfoGAN, InfoWGAN-GP (modified), VPGAN and Inference-InfoGAN (DCT)). 

\begin{figure*}[htb]
	\centering
	\includegraphics[width=1.0\textwidth]{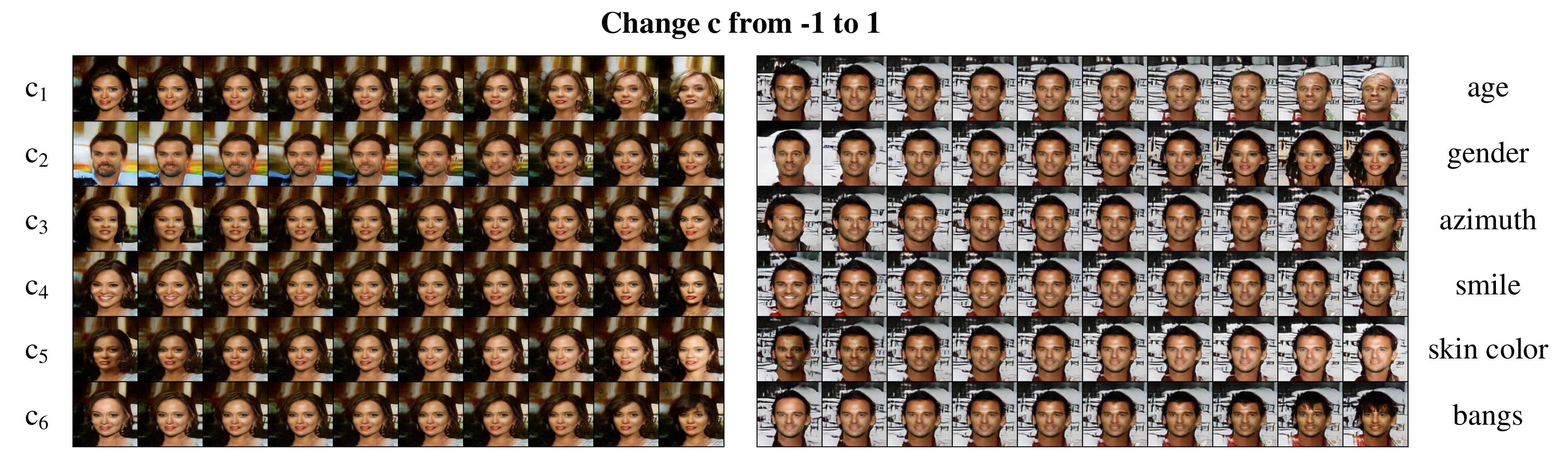}
	\caption{\textbf{Latent traversals trained on CelebA}. For the input $15$-dimensional $\bm{c}$ we select the $6$ dimensions with the most obvious disentanglement effects. They can obviously control the skin color, gender and other characteristics of the generated face images.}
	\label{face tra}
\end{figure*}

In terms of qualitative evaluation, it can be seen from Figure \ref{face tra} that our method obtains the latent variables corresponding to the skin color, gender and other characteristics of the generated face images. Figure \ref{fig:figure dsprites}(a) and Figure \ref{face tra} jointly illustrate that our proposed OBE module could infer target-wise interpretable latent variables via an adaptive orthogonal basis and further represents the unlabeled data, which mines richer information. 

In terms of quantitative evaluation, as shown in Table \ref{celeba scores}, our model achieves the best results in both VP and FID compared with other methods. Specifically, albeit slight improvement over FactorVAE in VP, the proposed Inference-InfoGAN significantly surpass FactorVAE in image FID, which reduces FID from $73.9$ to $37.9$. This indicates that the images generated by our model have higher quality. Also, comparisons with InfoGAN and InfoWGAN-GP validate the effectiveness of the proposed model, where the OBE module contributes to increments of $0.508$ and $0.106$ in VP and decrements of $5.4$ and $3$ in FID respectively. In comparison with VPGAN, the state-of-the-art method of considering VP and FID, our model achieves improvement of $0.093$ in VP and $9.5$ in FID. To summarize, our method generates higher-quility images than VAE-based models and shows the more independent latent variables compared with GAN-based models, i.e. guarantees both image quality and disentanglement ability.

\begin{figure*}[htb]
		\centering
		\includegraphics[width=1.0\textwidth]{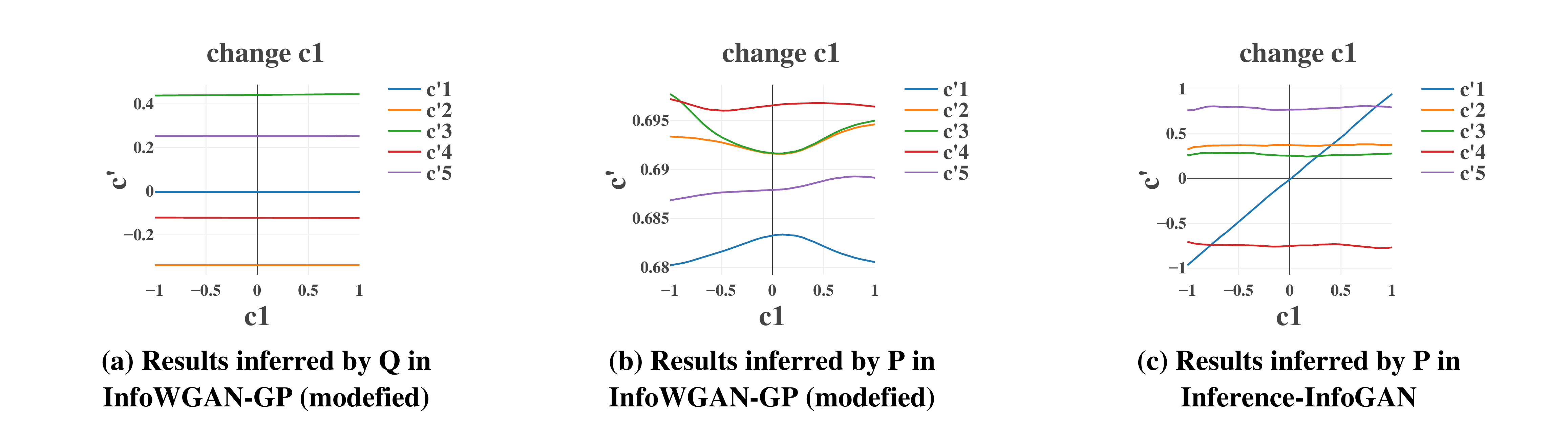}
		\caption{Correlation Curve of ${\bm{c}}$ and ${\bm{c}'}$ (On the trained basis)}
	\label{training c obe}
\end{figure*}

\begin{table}[htb]
    \caption{\textbf{Disentanglement and quality scores on CelebA dataset}. The smaller the quality score FID, the clearer the images generated by the model, the lowest is $0$. 
}
	\centering
	\setlength{\tabcolsep}{1.3mm}{
		\linespread{1.1}\selectfont
		\begin{tabular}{ccc}  \toprule
			Model  &VP &FID \\  \hline
			FactorVAE  &$0.750^\ddagger$ &$\textcolor[rgb]{1,0,0}{73.9 }^\ddagger$ \\
			InfoGAN  &$0.253^\ddagger$ &$43.3^\ddagger$ \\
			InfoWGAN-GP (modified)  &$0.651$ &$40.9$ \\
			VPGAN  &$0.668^\ddagger$ &$47.4^\ddagger$ \\
			\hline
			Inference-InfoGAN (DCT)  &$\textbf{0.757}$ &$46.3$ \\
			Inference-InfoGAN  &$\textbf{0.761}$ &$\textbf{37.9}$ \\
			\hline
	\end{tabular}}
    \label{celeba scores}
\end{table}

\subsection{Ablation studies}
\label{Ablation studies}

In this section, we perform experiments on dSprites to study the effects of our proposed OBE module, related hyperparameters, and alternating optimization technique on disentanglement.

\subsubsection{OBE module}

To analyze the effects of OBE component, we observe the relationship between the latent variables $\bm{c}$ and inference results on dSprites datasets. As shown in Figure \ref{training c obe}, one-dimensional $\bm{c}$ after adding the OBE module can control the change of the inference results based on a specific basis, but has less effects on others. In fact, when the basis of DCT is selected, the above independent influence is not obvious enough (seeing Appendix D). Therefore, training an orthogonal basis adaptively achieves better performance. Furthermore, whether it is based on encoder $Q$ of InfoGAN or adaptive orthogonal matrix ${\bm{P}}$ of our Inference-InfoGAN to infer the latent variables $\bm{c}$ in InfoGAN, the independence is not exhibited. It shows that the InfoGAN does not emphasize the inter-independence of latent variables, but does not mean that our orthogonal matrix cannot express the independent characteristics. According to the change of the inference results based on the proposed OBE module in our model, it can be seen that the latent variables are inter-independent.

In addition, to further reflect the importance of OBE, we conduct experiments under different hyperparameter $\lambda$, which could control the weight of introduced loss $\mathcal{L}_{inter-info}$ and experimental results has been given in Table \ref{tab:Disentanglement scores dsprites}. It can be seen that a greater $\lambda$ will lead to a higher and more stable average disentanglement score. Similar to the theoretical analysis in the method section, the results also illustrate that OBE component has a clear tendency to improve disentanglement. In general, emphasizing the constraints based on the OBE module will help enhance the inter-independence of the latent variables ${\bm{c}}$ and their interpretability in images ${\bm{X}}$, making up for the lack of independent representation of vanilla InfoGAN.

\begin{table}[htb]
    \caption{\textbf{Alternating optimization}. Alternating optimization brings a slight improvement on all the metrics.}
	\centering
	\setlength{\tabcolsep}{1.3mm}{
		\linespread{1.1}\selectfont
		\begin{tabular}{cccc}  \toprule
			  &FactorVAE &SAP &MIG \\  \hline
			w/o  &$0.930$ &$0.598$ 	&$0.373$ \\
			w  &$0.946$ &$0.622$ 	&$0.408$ \\
			\hline
	\end{tabular}}
    \label{Alternating optimization}
\end{table}

\subsubsection{Alternating optimization}

Alternating optimization is a type of training technique. In our design, it is firstly adopted to alternately train ${\bm{P}}$ alone for orthogonal regularization and optimization of the entire network parameters. To explore its effect, we compare the models obtained by alternating optimization and one-step optimization as shown in Table \ref{Alternating optimization}. The results
show that alternating optimization is beneficial for the training of our model to ensure the orthogonality of P and enhance disentanglement performance.

\section{Conclusion}
\label{conclusion}

We presented an approach for unsupervised disentanglement learning via embedding Orthogonal Basis Expansion on images. We prove that our method has a good trade-off between inter-independence inference ability and target-wise interpretability which are consistent with the disentanglement task. Experimentally, our work is better than the-state-of-art researches on different datasets in terms of multiple disentanglement metrics. Our method does not modify the hyperparameters when migrating from a labeled dataset to an unlabeled dataset, but the selection of the hyperparameter is still a key step to further improve the disentanglement performance of the model. Therefore, in future work, we will continue to study unsupervised hyperparameter adjustment and other limitations of our work, such as experiments on more difficult datasets. 

\section{Appendix}

\subsection{A. Derivation of variational lower bound}

\label{app:Derivation of variational lower bound}

\begin{equation}\tag{14}
\label{variational lower bounds}
\begin{aligned}
    & \mathcal{L}_{infer-info}(G,Q,\bm{P}) \\=&\mathbb{E}_{\bm{c}\sim p(\bm{c}),\bm{X}\sim p(G(\bm{z},\bm{c}))}\bigg[\lambda\log{\prod_iq'(c_i|c'_i;\bm{P})}
    \\&+(1-\lambda)\log{q(\bm{c}|\bm{X};Q)}\bigg]\\
    =&\mathbb{E}_{\bm{c}\sim p(\bm{c}),\bm{X}\sim p(G(\bm{z},\bm{c}))}\bigg[
    -\lambda\log{\frac{q(\bm{c}|\bm{X};Q)}
    {\prod_iq'(c_i|c'_i;\bm{P})}}\\&+\log{q(\bm{c}|\bm{X};Q)}\bigg]\\
    =&-\lambda\mathbb{E}_{\bm{X}\sim p(G(\bm{z},\bm{c}))}\left[D_{KL}\left(p(\bm{c}|\bm{X})||\prod_i q'(c_i|\bm{X};\bm{P})\right)\right]\\
    &+\mathbb{E}_{\bm{c}\sim p(\bm{c}),\bm{X}\sim p(G(\bm{z},\bm{c}))}\left[
    \lambda\log{\frac{
    p(\bm{c}|\bm{X})}
    {q(\bm{c}|\bm{X};Q)}}+\log{q(\bm{c}|\bm{X};Q)}\right]\\
    =&-\lambda\mathbb{E}_{\bm{X}\sim p(G(\bm{z},\bm{c}))}\left[D_{KL}\left(p(\bm{c}|\bm{X})||\prod_i q'(c_i|\bm{X};\bm{P})\right)\right]\\
    &+(\lambda-1)\mathbb{E}_{\bm{X}\sim p(G(\bm{z},\bm{c}))}\bigg[D_{KL}\left(p(\bm{c}|\bm{X}) || q(\bm{c}|\bm{X};Q)\right)\bigg]\\
    &+{I}(\bm{c};G(\bm{z},\bm{c}))-H(\bm{c})\\
    \leq&-\lambda\mathbb{E}_{\bm{X}\sim p(G(\bm{z},\bm{c}))}\left[D_{KL}\left(p(\bm{c}|\bm{X})||\prod_i q'(c_i|\bm{X};\bm{P})\right)\right]\\&+{I}(\bm{c};G(\bm{z},\bm{c}))-H(\bm{c}),
\end{aligned}
\end{equation}

\subsection{B. Algorithm}

\begin{algorithm}
	\caption{Inference-InfoGAN}
	\label{alg}
	\begin{algorithmic}[1]
	    \REQUIRE initial parameters ${\theta}_D$, ${\theta}_G$, ${\theta}_Q$, ${\theta}_P$ for $D$, $G$, $Q$, $P$, training iterations $n$ and other hyperparameters
	    \FOR{$i \gets 1$ to $n$}
		\STATE Sample $\bm{\mu}$, images $\bm{X}$, noise $\bm{z}$, latent variables $\bm{c}$.
		\STATE $\tilde{\bm{X}} \gets G(\bm{z},\bm{c})$
		\STATE $\hat{\bm{X}}_j \gets \bm{\mu}_j\bm{X}+(1-\bm{\mu}_j)\tilde{\bm{X}}_j$
		\STATE Update ${\theta}_D$ based on Equation~(\ref{eq:WGAN-div loss}).
		\STATE Compute $\mathcal{L}$ based on Equation~(\ref{eq:WGAN-div loss}) and Equation~(\ref{eq:inter-info}): 
$\mathcal{L} \gets \mathcal{L}_{inter-info}-\mathbb{E}_{{\bm{z}}\sim p_{z}({\bm{z}})}\left[D(G({\bm{z}}))\right]$
		\STATE Update ${\theta}_G$, ${\theta}_Q$, ${\theta}_P$ based on $\mathcal{L}$.
		\WHILE{$\mathcal{L}_{or} >= \epsilon$}
		\STATE Update ${\theta}_P$ based on Equation~(\ref{eq:Lor}).
		\ENDWHILE
		\ENDFOR
	\end{algorithmic}  
\end{algorithm}

\subsection{C. Choice of orthogonal basis}
\label{app:Choice of orthogonal basis}
In our proposed OBE module, different from training an orthogonal matrix $\bm{P}$ to obtain one set of orthogonal basis, another simple method is to directly select meaningful and interpretable orthogonal basis. Therefore, we also try to use the basis of Discrete Cosine Transform (DCT) on the experiment. According to the Inverse Discrete Fourier Transform, an image $\bm{I}\in{\mathbb{R}^{n\times{n}}}$ can be expanded as follows:

\begin{equation}\tag{15}
\label{eq:IDFT}
\begin{aligned}
{I}_{xy}=\frac{1}{N^2}\sum_{u=0}^{N-1}\sum_{v=0}^{N-1}c'_{uv}e^{j2\pi(\frac{ux}{N}+\frac{vy}{N})}.
\end{aligned}
\end{equation}

The Fourier basis is naturally orthogonal, so the above formula can be understood as expanding the image on an orthogonal basis, and $c'_{uv}$ can represent the coefficients corresponding to different orthogonal basis:

\begin{equation}\tag{16}
\label{eq:DFT}
\begin{aligned}
c'_{uv}=\sum_{x=0}^{N-1}\sum_{y=0}^{N-1}I_{xy}e^{-j2\pi(\frac{ux}{N}+\frac{vy}{N})}.
\end{aligned}
\end{equation}

In order to simplify the calculation, we choose DCT that ignores the imaginary part. And the DCT coefficients are modified to ensure the orthogonality of the basis, specifically as follows:

\begin{equation}\tag{17}
\label{eq:DCT}
\begin{aligned}
c'_{uv}=&\frac{1}{N}f(u)f(v)\sum_{x=0}^{N-1}\sum_{y=0}^{N-1}I_{xy}\\
*&\;cos\left[\frac{(x+0.5)\pi}{N}u\right]cos\left[\frac{(y+0.5)\pi}{N}u\right],
\end{aligned}
\end{equation}
where
\begin{align}\tag{18}
f(u)=\left\{
\begin{aligned}
1, u=0 \\
2, u\neq0
\end{aligned}
\right.
\end{align}

According to Eq .\ref{eq:DCT}, the optimization of the objective function $\mathcal{L}_{infer-info}(G,Q,\bm{P})$ can be completed. Furthermore, we considered the physical meaning of the DCT basis. It has a specific meaning to the image, one part of which represents low-frequency signals, and the other part represents high-frequency signals of the images. We assume that the main information of the image comes from $\bm{z}$, which corresponds to low-frequency signals, while the latent variables $\bm{c}$ supplement the details of the images corresponding to high-frequency signals. Therefore, we choose the ${k}$ DCT coefficients corresponding to high-frequency and ${k}$-dimensional vector $\bm{c}$ to calculate $\mathcal{L}_{infer-info}$ when implementing the experiment.

\subsection{D. Inference independence based on DCT}

\begin{figure}[H]
		\centering
		\includegraphics[width=\columnwidth]{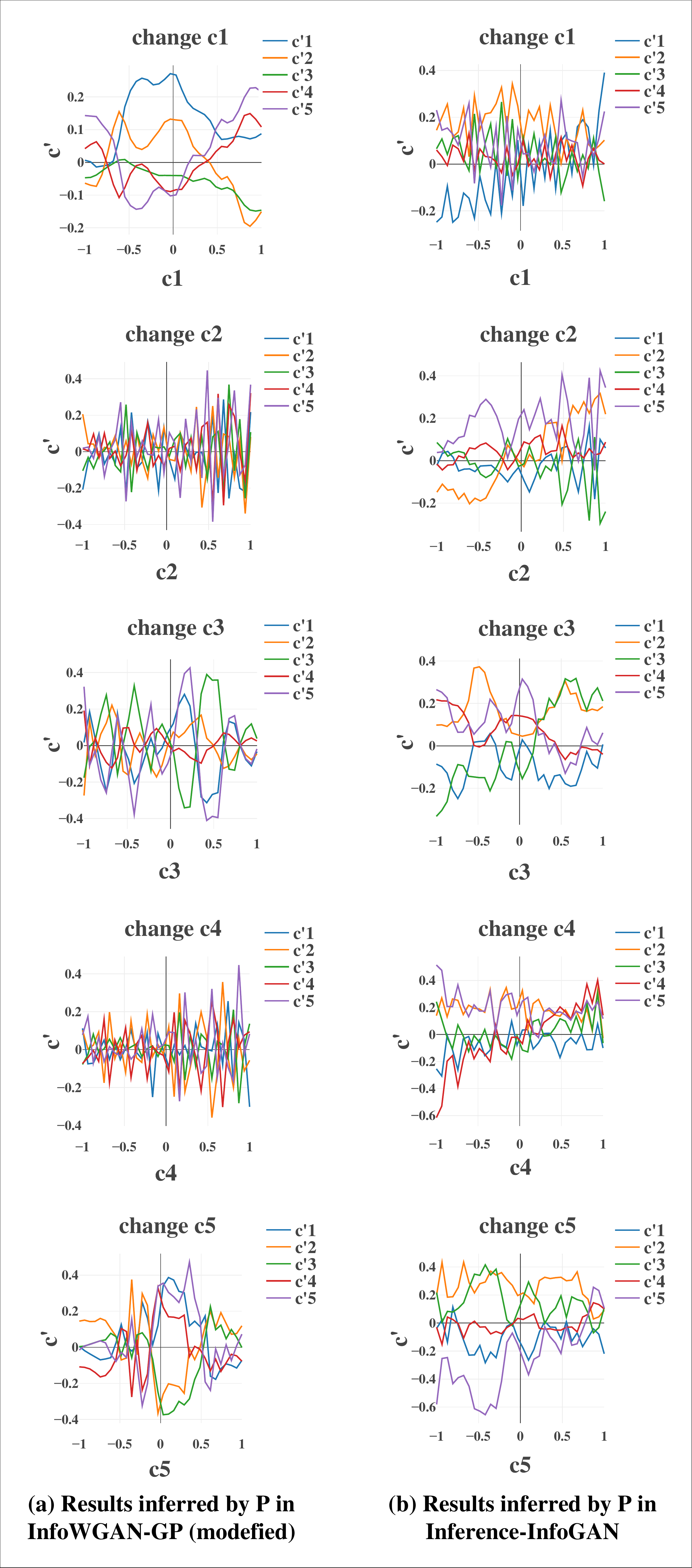}
		\caption{Correlation Curve of ${\bm{c}}$ and ${\bm{c}'}$ (On the basis of DCT)}
		\label{dct c obe}
\end{figure}

We follow the method in Appendix C to obtain $\bm{P}$ and perform ablation analysis on the OBE module. The results are shown in Figure \ref{dct c obe}. 

Compared with adaptively training one set of orthogonal basis, the independent influence of the latent variables $\bm{c}$ on the OBE inference results obtained by directly selecting the basis of DCT is significantly weaker. This shows that using the basis of DCT to restrain the independence of latent variables and making latent variables control the interpretable representation is a contradictory process. So even if it can be disentangled eventually, changing a single-dimension latent variable will still have an impact on other dimensions. Therefore, our proposed adaptive OBE can express better independent characteristics than fixed basis expression of DCT.

\bibliography{citation}

\end{document}